\AtBeginDocument{%
  \providecommand\BibTeX{{%
    \normalfont B\kern-0.5em{\scshape i\kern-0.25em b}\kern-0.8em\TeX}}}
\documentclass[sigconf]{acmart}
\usepackage{algorithm,float}
\usepackage{algorithmic}
\usepackage{amsmath}
\usepackage{graphicx}
\usepackage{subfigure}
\usepackage{graphicx}
\usepackage{float}
\makeatletter
\usepackage{verbatim}
\makeatletter
\usepackage{balance}
\usepackage[british]{babel}
\usepackage{stfloats}
\usepackage{amsmath}
\graphicspath{{figs/}}
\newenvironment{breakablealgorithm}
  {
   \begin{center}
     \refstepcounter{algorithm}
     \hrule height.8pt depth0pt \kern2pt
     \renewcommand{\caption}[2][\relax]{
       {\raggedright\textbf{\ALG@name~\thealgorithm} ##2\par}%
       \ifx\relax##1\relax 
         \addcontentsline{loa}{algorithm}{\protect\numberline{\thealgorithm}##2}%
       \else 
         \addcontentsline{loa}{algorithm}{\protect\numberline{\thealgorithm}##1}%
       \fi
       \kern2pt\hrule\kern2pt
     }
  }{
     \kern2pt\hrule\relax
   \end{center}
  }
\makeatother
\usepackage{setspace}

\copyrightyear{2021} 
\acmYear{2021} 
\setcopyright{acmlicensed}\acmConference[KDD '21]{Proceedings of the 27th ACM SIGKDD Conference on Knowledge Discovery and Data Mining}{August 14--18, 2021}{Virtual Event, Singapore}
\acmBooktitle{Proceedings of the 27th ACM SIGKDD Conference on Knowledge Discovery and Data Mining (KDD '21), August 14--18, 2021, Virtual Event, Singapore}
\acmPrice{15.00}
\acmDOI{10.1145/3447548.3467204}
\acmISBN{978-1-4503-8332-5/21/08}
\begin{document}
\fancyhead{}
\title{Multi-Agent Cooperative Bidding Games for Multi-Objective Optimization in e-Commercial Sponsored Search}

\author{Ziyu Guan$^{1\dagger}$, Hongchang Wu$^{2,1\dagger}$, Qingyu Cao$^{2\ddagger}$, Hao Liu$^{2}$, Wei Zhao$^{1\ddagger}$}
\author{Sheng Li$^2$, Cai Xu$^1$, Guang Qiu$^2$, Jian Xu$^2$, Bo Zheng$^2$}
\affiliation{%
  \institution{$^1$Xidian University \country{China}  \qquad $^2$Alibaba Group \country{China} }}
\email{{zyguan@, ywzhao@mail., cxu@}xidian.edu.cn}
\email{{wuhongchang.whc, qingyu.cqy, liuhao.lh, junqian.ls, xiyu.xj, guang.qiug, bozheng}@alibaba-inc.com}

\thanks{$\dagger$: both authors contributed equally to this research. $\ddagger$: corresponding author.}

\renewcommand{\shortauthors}{Guan and Wu, et al.}
\begin{abstract}

Bid optimization for online advertising from single advertiser's perspective has been thoroughly investigated in both academic research and industrial practice. However, existing work typically assume competitors do not change their bids, i.e., the wining price is fixed, leading to poor performance of the derived solution. Although a few studies use multi-agent reinforcement learning to set up a cooperative game, they still suffer the following drawbacks: (1) They fail to avoid collusion solutions where all the advertisers involved in an auction collude to bid an extremely low price on purpose. (2) Previous works cannot well handle the underlying complex bidding environment, leading to poor model convergence. This problem could be amplified when handling multiple objectives of advertisers which are practical demands but not considered by previous work. In this paper, we propose a novel multi-objective cooperative bid optimization formulation called \emph{Multi-Agent Cooperative bidding Games (MACG)}. MACG sets up a carefully designed multi-objective optimization framework where different objectives of advertisers are incorporated. A global objective to maximize the overall profit of all advertisements is added in order to encourage better cooperation and also to protect self-bidding advertisers. To avoid collusion, we also introduce an extra platform revenue constraint. We analyze the optimal functional form of the bidding formula theoretically and design a policy network accordingly to generate auction-level bids. Then we design an efficient multi-agent evolutionary strategy for model optimization. Evolutionary strategy does not need to model the underlying environment explicitly and is more suitable for bid optimization. Offline experiments and online A/B tests conducted on the Taobao platform indicate both single advertiser's objective and global profit have been significantly improved compared to state-of-art methods.

\end{abstract}

\keywords{E-commercial Sponsored Search, Multi-objective Optimization, Neural Evolutionary Strategies, Multi-Agent Cooperative Games}
\maketitle
\section{INTRODUCTION}
Online advertising \cite{goldfarb2011online,evans2009online} is a marketing strategy utilizing the Internet as a medium to help advertisers attract target audiences and conversions via \emph{Real-Time Biding (RTB)}. E-commercial sponsored search is a mainstream form of online advertising, for advertisers to promote their products or services. E-commercial sponsored search generally ranks candidate advertisements (ADs) by \emph{effective Cost Per Mille (eCPM)} (i.e., the product of the predicted \emph{Click-Through Rate (CTR)} and the advertiser's bid) and charges the winner(s) via \emph{Generalized Second Price (GSP)} once clicked. The ranking and deduction together defines the auction mechanism. It has been proved that in a single auction with single winner, GSP is incentive compatible that truthful bidding is the optimal choice for every participant \cite{wilkens2017gsp,cavallo2018matching}. However, in a sequential auction process with coupled constraints such as budget, truthful bidding often yields early exhaustion of budget and poor performance for advertisers. 


In the highly dynamic marketing environment of e-Commerce platforms, there are typically millions of advertisers, billions of \emph{Page-Views (PVs)} and hundreds of billions of auctions\footnote{We assume there is only one winner in each auction. For multiple winners in a single auction, we can treat it as multiple auctions with the same bidding.} every day. Bid optimization is an effective way to achieve customer objectives and has been investigated thoroughly. Most existing works \cite{cai2017real,borgs2007dynamics,evans2009online} optimize the total utility (be it clicks, conversions or \emph{Gross Merchandise Volume (GMV)}) of a single customer across a day, given the daily budget constraint and/or the \emph{Pay Per Click (PPC)} constraint, assuming competitors do not change their bids. Such separate optimization process often leads to sub-optimal solutions because in an auction competitors easily change the winning price of each other. Some recent works \cite{jin2018multi,zhao2018deep} try to solve this problem by setting up a multi-agent cooperative game via \emph{Reinforcement Learning (RL)}. Nevertheless, they still suffer the following drawbacks: (1) They fail to avoid collusion solutions where all the advertisers involved in an auction collude to bid an extremely low price on purpose. Such collusion solutions will not reduce the global profit of advertisers and easily satisfy the cost constraints. However, it deprives the motive of the platform to optimize for common good. Although self-bidding advertisers help alleviate this issue, it is non-negligible as more and more advertisers opt in bid optimization. (2) They cannot well handle the underlying complex bidding environment, leading to poor model convergence. Specifically, the underlying transition mechanisms could be very different between, e.g., weekdays and weekends (or different seasons). This problem could be amplified exponentially when handling multiple objectives of advertisers which are practical demands but not considered by previous work.

The goal of bid optimization is to satisfy different demands of the advertisers, as well as to keep the ecosystem healthy. In light of this, building cooperative bidding agents with different objectives of advertisers is challenging. Firstly, cooperation in an auction might leads to the collusion solutions discussed above. Secondly, optimizing the bidding strategies of multiple ADs simultaneously inflate the solution space exponentially. For example, given the possible $B$ bid choices, $M$ ADs and $N$ auctions, the possible solution space is in the scale of $\mathcal{O}(B^{MN})$. Thirdly, how to incorporate different advertisers' objectives together is also crucial. These advertisers might include not only those who opt in bid optimization but also those who do not. For those who already opted in, it motivates them to willingly keep using the bid optimization service to let the game roll on. For the remaining self-bidding advertisers, their profit must also be taken care of in an indirect way such that they can also acquire appropriate PVs.

In this paper, we explicitly model the multi-AD optimization problem in a novel multi-objective cooperative bid optimization formulation called Multi-Agent Cooperative bidding Games (MACG). ADs equipped with different private objectives are optimized together. To avoid trivial collusion solutions, we introduce an additional platform revenue constraint, i.e., the \emph{Revenue Per Mille (RPM)} constraint. In addition to smart-bidding advertisers, we also set the GMV of all advertisers as a global objective, to avoid extreme low utility of self-bidding advertisers. Furthermore, we establish an explicit lower bound on the utility achieved by smart-bidding advertisers, to keep them motivated. We give a theoretical analysis of the MACG problem to discuss the optimal functional form of the bidding formula. A policy net is then designed based on this optimal bidding form. The policy net sets up an agent net for each objective to generate real-time bids for all the ADs with that objective (agent sharing helps save parameter space and avoid overfitting). The generated bids are further adjusted by the output of a shared net which aims to achieve the global objective. We also design an efficient multi-objective evolutionary strategy (ES) to optimize network parameters for large-scale industrial environments. Since ES does not need to model the underlying environment explicitly, it would not suffer the aforementioned problem of RL.

The major contributions can be summarized as follows: (1) We propose a novel multi-AD cooperative games formulation for bid optimization in e-commercial sponsored search. MACG considers not only multi-objectives of smart-bidding advertisers but also the benefits of self-bidding advertisers and the platform. (2) We theoretically analyze the optimal functional bidding formula under MACG. A multi-agent policy net is designed based on the optimal bidding form to solve MACG, encoding both selfish and global objectives. In order to achieve timely training of the model for large-scale industrial environments, we heuristically design an efficient multi-agent ES to update network parameters. In each iteration, it simply tries to move towards Pareto optimal solutions, i.e., trying to optimize the lowest objective together with the global objective, without degenerating the other objectives. (3) For empirically evaluation, we deploy MACG in a distributed computation environment in Taobao's search auction platform which possesses billions of online auctions and millions of active ADs every day. Offline evaluation and standard online A/B tests prove the superiority of MACG compared to state-of-art methods.



\section{RELATED WORK}
\subsection{Bid optimization}
Existing work for bid optimization can mainly be divided into two categories: static optimization methods assuming that the other advertisers’ bids will not change, and dynamic optimization methods taking into account the coupling of advertisers. As for static optimization methods, Perlich \textit{et al.} \cite{perlich2012bid} introduced a linear bidding strategy; Zhang \textit{et al.} \cite{zhang2014optimal} explored the non-linear relationship between optimal bid and auction evaluation. Kitts \textit{et al.} \cite{kitts2004optimal} focused on keywords' utility estimation in sponsored search RTB. Zhu \textit{et al.} \cite{zhu2017optimized} proposed the \emph{Optimized Cost Per Click (OCPC)} bidding algorithm for display advertising, which has been widely used in real-world applications. More recent bid optimization algorithms modeled the sequential auction process in the \emph{Markov Decision Process (MDP)} framework and took RL to optimize bidding strategy. Cai \textit{et al.} \cite{cai2017real} and Zhao \textit{et al.} \cite{zhao2018deep} utilized RL to learn the optimal bid for a single AD in display advertising and sponsored search, respectively. These methods learned the optimal bidding strategy for each AD separately, and therefore failed to consider bidding interactions in the complex auction environment.

As for dynamic optimization, Zhao \textit{et al.} \cite{zhao2018deep} designed a massive-agent RL model which used a global cooperative objective to incorporate agents' interactions. However, it is still difficult to be applied to millions of ADs due to prohibitive computational complexity of training one model for each AD and data sparsity for ``long tail'' ADs (i.e., no sufficient data for training their separate agent models). Jin \textit{et al.} \cite{jin2018real} clustered millions of ADs by layered daily revenue to reduce parameter space and proposed a practical Distributed Coordinated Multi-Agent Bidding (DCMAB) framework. However, DCMAB (and also \cite{zhao2018deep}) failed to avoid collusion solutions which are harmful for the platform. Furthermore, since RL cannot well handle the complex bidding environment, they are difficult to converge even for the single-objective case, not to mention the more complicated multi-objective case.

\subsection{Multi-objective Optimization}
The goal of a \emph{Multi-Objective Optimization (MOOP)} problem is generally to find the Pareto optimal solutions. A solution is said to dominate another one if for every objective the utility is no less. A Pareto optimal solution is the solution which cannot be dominated by other solutions. All Pareto optimal solutions constitute the Pareto optimal solution set. Existing solutions to MOOP can be divided into three categories by how they deal with different objectives. The first category of methods prioritizes different objectives according to corresponding constraints and optimizes each objective from high priority to low priority, e.g., \cite{geyik2016joint}. The second one is to convert multiple objective functions into a single objective function through linearly weighted summation with non-negative weights, and then apply single-objective optimization techniques e.g., \cite{zitzler1998multiobjective}. The third category consists of Multi-objective evolutionary algorithms \cite{ribeiro2014multiobjective,zitzler1998multiobjective}. For MACG, there is no priority trade-off between different objectives; linear combination of objectives requires searching for proper weights, which is time-consuming. Hence, we choose to design an efficient multi-objective evolutionary algorithm for optimization. Evolutionary algorithms can naturally explore a balance between the cooperation and competition among advertisers' multiple objectives and the global objective.

In \cite{wang2012multi}, a Multi-Objective Optimization (MOO) framework is proposed to optimize the profit of platform, advertisers and query users together by tuning the ranking function (i.e., auction mechanism). Our work is orthogonal to theirs in that we aim to achieve different objectives of advertisers and keep auctions healthy by optimizing their bidding policies while keeping the auction mechanism fixed. MOO is not directly applicable to our problem.

\subsection{Evolutionary Strategies}
\emph{Evolution Strategy (ES)} \cite{huning1976evolutionsstrategie} is considered as a particular set of optimization algorithms which has the properties of no need for backpropagating gradients and tolerance of potentially arbitrarily long time horizons. In each iteration of ES, a set of parameter vectors will be generated, and their objective function values will be evaluated. The parameter vectors with the highest scores are then reorganized to form the next-generation population, and this process is repeated until the objective converges. Covariance matrix adaptation evolution strategy (CMA-ES) \cite{iruthayarajan2010covariance, hansen2001completely} and Natural Evolution Strategies (NES) \cite{sun2009efficient,sehnke2010parameter} are two most widely known memebers of ES for the single objective case. In order to solve multi-objective problems, Zitzler \textit{et al.} \cite{zitzler1998multiobjective} proposed a non-dominated sorting genetic algorithm which suffered computational complexity of $O(M*N^3)$, where $M$ is the number of objectives and $N$ is the population size. Afterwards, the non-dominated sorting genetic algorithm $\uppercase\expandafter{\romannumeral2}$ ($NSGA\uppercase\expandafter{\romannumeral2}$) was proposed in \cite{deb2002fast}. It reduced the computational complexity to $O(M*N^2)$. In large-scale industrial bidding environments, the complexity of ($NSGA\uppercase\expandafter{\romannumeral2}$) is still high. In this paper we heuristically design a more flexible and lightweight multi-objective ES with $O(N)$ to optimize MACG.

\section{PROBLEM DEFINITION}
\begin{table}[]
\caption{Common notations}
    \centering
    \begin{tabular}{c|c}
     \hline
           Notations & Description\\
            \hline
            $i$ / $j$ & an AD/auction \\
            $\mathcal{I}$ / $\overline{\mathcal{I}}$ & the set of smart-bidding/self-bidding ADs \\
            $\mathcal{J}$ & the set of auctions in a day \\
            $f_i(\cdot)$ & objective of AD $i$, $\forall i \in \mathcal{I}$ \\
            $f_0(\cdot)$ & global GMV objective \\
            $k$ & (index of) an formulated objective \\ 
            $\mathcal{I}^{(k)}$ & the set of ADs with objective $k$\\
            $\mathcal{I}_{j}$ &  the set of ADs under auction $j$\\
            $\mathcal{I}_{j}^{(k)}$ & the set of ADs with objective $k$ under auction $j$\\
            $K$ & the number of different objectives \\
            $e_{ij}$ & indicator: whether AD $i$ win auction $j$ \\ 
            $b_{ij}$ & the generated bid of AD $i$ for auction $j$ \\
            $w_{ij}$ & the winning price of AD $i$ for auction $j$ \\
            $g_k(i,j)$ & the objective function in type $k$ of AD $i$ for auction $j$ \\ 
            $\mathbf{{S_{ij}}}$ & the feature vector of AD $i$ under auction $j$\\
            $\mathbf{{S_{.j}}}$ & the summarized feature vector under auction $j$\\
            $M^{(k)}$ / $M^{(0)}$ & Objective score of objective $k$/global objective \\
          \hline
    \end{tabular}
    \label{tab:notations}
\end{table}
In this section, we mathematically formulate the problem of MACG in sponsored search. We consider MACG as an episodic bidding process with auction set $\mathcal{J}$ across a day. Let $\mathcal{I}$ and $\overline{\mathcal{I}}$ be the sets of smart-bidding advertisers and self-bidding advertisers, respectively. Without loss of generality, advertisers can be replaced with their advertisements (ADs), as long as the objective and constraints hold upon an AD. Accordingly, we use $i \in \mathcal{I} \cup \overline{\mathcal{I}}$ and $j \in \mathcal{J}$ to represent an AD and an auction, respectively. $\mathcal{I}$ can be grouped into different clusters according to their objectives, i.e., $\mathcal{I} = \mathcal{I}^{(1)} \cup \dots \cup \mathcal{I}^{(K)}$, where $\mathcal{I}^{(k)} \cap \mathcal{I}^{(k')} = \emptyset$, $\forall k \neq k'$. We use $\mathcal{I}_{j}$ to denote the set of ADs involved in auction $j$, and $\mathcal{I}_{j}^{(k)} = \mathcal{I}_{j} \cap \mathcal{I}^{(k)}$ to denote the set of ADs with objective $k$ under auction $j$. The generated bid of AD $i$ for auction $j$ is denoted by $b_{ij}$. Under each auction $j$, the ranking score of an AD $i \in \mathcal{I}_{j}$ in typical GSP is calculated by $CTR_{ij} * b_{ij}$ (i.e., eCPM) and the top ranked AD winning $j$ will be presented to the user. Here, $CTR_{ij}$ is the
predicted click-through rate of the user-query-item triple \cite{zhou2018deep,zhou2019deep}. In this paper, our focus is not on how to estimate CTR and we treat it as known constant here. We also define a set of indicator variables $\{e_{ij}\}$:
\begin{equation}
e_{ij} = 
\begin{cases}
1, \quad & \textrm{if AD } i \textrm{ wins auction } j, \\
0, \quad & \textrm{otherwise}.
\end{cases}
\end{equation}
For the typical GSP mechanism \cite{wilkens2017gsp} in sponsored search, the expected cost $w_{ij}$ for AD $i$ to win auction $j$ is:
\begin{equation}
w_{ij} = CTR_{ij} * \frac{CTR_{next} * b_{next}}{CTR_{ij}} = CTR_{next} * b_{next},
\end{equation}
where $CTR_{next}$ and $b_{next}$ are the estimated CTR and bid for the next ranking position according to eCPM. From the definitions of $\{e_{ij}\}$ and $\{w_{ij}\}$, we can see that they are both determined by $\{b_{ij}\}$. 
Different from the single AD case, in MACG $\{w_{ij}\}$ cannot be treated as constant and it is one of the major difficulties faced here.


The objective of AD $i \in \mathcal{I}^{(k)}$ is calculated as follows
\begin{equation*}
    f_i(\cdot) = \sum\limits_{j \in \mathcal{J}}{e_{ij} g_k(i,j)}, \quad \forall i \in \mathcal{I}^{(k)}
\end{equation*}
Typical objectives of $g_k(i,j)$ and their calculations in sponsored search are summarized in Tab. \ref{tab:objectives}. $CVR_{ij}$, $IP_{ij}$, $WCVR_{ij}$ represent predicted conversion rate, item price, predicted weak conversion rate (i.e., favorite or adding to shopping cart) for AD $i$ under auction $j$, respectively. Similar with $CTR_{ij}$, we treat $CVR_{ij}$ and $WCVR_{ij}$ as known variables obtained from an oracle.
\begin{table}[]
\caption{Objectives and calculations}
    \centering
    \begin{tabular}{c|c}
     \hline
           Objectives & Calculation \\
            \hline
            $CLK_{ij}$ & $CTR_{ij}$ \\
            \hline
            $CART_{ij}$ & $CTR_{ij} \cdot WCVR_{ij}$ \\ 
            \hline
            $GMV_{ij}$ & $CTR_{ij} \cdot CVR_{ij} \cdot IP_{ij}$ \\ 
          \hline
    \end{tabular}
    \label{tab:objectives}
\end{table}


Besides single AD's objective, we also include an additional $f_0(\cdot)$ in the objective. It represents the global GMV of the whole AD set $\mathcal{I} \cup \overline{\mathcal{I}} $ earned from auction set $\mathcal{J}$. Such explicit global profit emphasizes the importance of purchases in a performance-based advertising platform. Furthermore, we also encode self-bidding advertisers' GMV in it, to encourage explicit cooperation. $f_0(\cdot)$ can be calculated as follows.
\begin{equation*}
    f_0(\cdot) = \sum\limits_{i \in {\mathcal{I} \cup \overline{\mathcal{I}}}}{\sum\limits_{j \in \mathcal{J}}{e_{ij} \cdot g_0(i,j)}},
\end{equation*}
where $g_0(i,j) = CTR_{ij} \cdot CVR_{ij} \cdot IP_{ij} = GMV_{ij}$. 


For constraints, each AD has its own budget constraint $B_i$ and PPC constraint $C_i$ set by advertisers (if not, $B_i$ and $C_i$ are infinite). The additional lower bound of platform revenue constraint (RPM constraint) to avoid collusion is denoted as $\eta$, which can be interpreted as a reasonable take-rate of the global GMV. We use the dynamic lower bound to constrain platform revenue to avoid the ambiguity and risk in determining a fixed lower bound. Furthermore, to motivate the smart-bidding advertisers to keep in the game, we introduce a lower bound for $f_i(\cdot)$ as $F_i$ for each AD $i$. $F_i$ is reasonable only when it is no less than self-bidding and can be estimated from our historical auction logs. We give the complete problem formulation as:
\begin{equation}\label{eqn:primal}
    \begin{aligned}
        \max_{\{b_{ij}\}} \quad & \mathbf{F}(e_{ij}) = [f_0(\cdot), f_1(\cdot), ..., f_{|\mathcal{I}|}(\cdot)]^T \\
        s.t. \quad & \sum\limits_{j \in \mathcal{J}}{e_{ij} \cdot w_{ij}} \le B_i, \forall i \in \mathcal{I}, \\
        \quad & \frac{\sum\limits_{j \in \mathcal{J}}{e_{ij} \cdot w_{ij}}}{\sum\limits_{j \in \mathcal{J}}{e_{ij} \cdot CTR_{ij}}} \le C_i, \forall i \in \mathcal{I}, \\
        \quad & f_i(\cdot) \ge F_i, \forall i \in \mathcal{I}, \\
        \quad & \sum\limits_{i \in \mathcal{I} \cup \overline{\mathcal{I}}}{\sum\limits_{j \in \mathcal{J}}{e_{ij} \cdot w_{ij}}} \ge \eta f_0(\cdot), \\
        \quad & \sum\limits_{i \in \mathcal{I} \cup \overline{\mathcal{I}}}{e_{ij}} \le 1, \forall j \in \mathcal{J}.
    \end{aligned}
\end{equation}
It should be pointed out that the set $\{e_{ij}, \forall i \in \mathcal{I} \cup \overline{\mathcal{I}}\}$ is determined by the bidding set $\{b_{ij}, \forall i \in \mathcal{I} \cup \overline{\mathcal{I}}\}$, where the latter is only partially controlled by the platform (only the smart-bidding ADs). Note that we assume that there is only one winner in each auction in the final constraint (it is an inequality since there could be no winner due to out of budget). Common notations defined above are summarized in Tab. \ref{tab:notations} for clarity.

\section{Theoretical Analysis}
In Eq.~(\ref{eqn:primal}) we can see that there are two kinds of objectives, the global objective $f_0(\cdot)$ and private objectives $\{f_i(\cdot)\}$. The two kinds of objectives are constrained by the coupled RPM constraint and the assignment constraint. Fortunately, the two constraints are linear with respect to the assignment variables $\{e_{ij}\}$.

Considering the following sub-problem for the global objective:
\begin{equation}\label{eqn:sub_primal1}
\begin{aligned}
    \max_{\{b_{ij}\}} \quad & f_0(\cdot) \\
    s.t. \quad & \sum\limits_{i \in \mathcal{I} \cup \overline{\mathcal{I}}}{\sum\limits_{j \in \mathcal{J}}{e_{ij} \cdot w_{ij}}} \ge \eta f_0(\cdot), \\
    \quad & \sum\limits_{i \in \mathcal{I} \cup \overline{\mathcal{I}}}{e_{ij}} \le 1, \forall j \in \mathcal{J}.
\end{aligned}
\end{equation}
For a single auction, it is obvious that if each AD bids as:
\begin{equation*}
    b_{ij} = \alpha_j \cdot CVR_{ij} \cdot IP_{ij}, \forall i \in \mathcal{I}, j \in \mathcal{J},
\end{equation*}
then all the ADs are ranked based on $\alpha_j * GMV_{ij}$ (eCPM). In this sense, we can assure $f_0(\cdot)$ over auction $j$ is maximized even we do not control all the bids. Considering the RPM constraint, letting the AD with the largest $GMV$ win the auction also makes it easier to satisfy the RPM constraint since we can control the second price via smart-bidding advertisers. The optimal value of $f_0(\cdot)$ depends on the specific value of $\alpha_j$. And $\alpha_j$ is determined by the linear constraints. We argue that the optimal functional form preserves. 

For each private $f_i(\cdot)$, the related sub-problem is (we omit constraint $f_i(\cdot) \ge F_i$ since it does not affect the optimal solution form):
\begin{equation}\label{eqn:sub_primal2}
\begin{aligned}
\max_{\{b_{ij}\}} \quad & f_i(\cdot) \\
s.t. \quad & \sum\limits_{j \in \mathcal{J}}{e_{ij} \cdot w_{ij}} \le B_i, \\
\quad & \frac{\sum\limits_{j \in \mathcal{J}}{e_{ij} \cdot w_{ij}}}{\sum\limits_{j \in \mathcal{J}}{e_{ij} \cdot CTR_{ij}}} \le C_i, \\
\quad & e_{ij} \le 1, \forall j \in \mathcal{J}.
\end{aligned}
\end{equation}
If all the competitors stay unchanged, the optimal bidding formula for $f_i(\cdot)$ can be calculated as \cite{yang2019bid}:
\begin{equation*}
    b_{ij} = \beta_i \cdot \frac{g_k(i,j)}{CTR_{ij}} + \gamma_i C_i, \forall i \in \mathcal{I},
\end{equation*}
where the first term optimizes the private objective and the second term tackles the PPC constraint. The parameter pair $(\beta_i, \gamma_i)$ is determined by the auction distribution, winning price distribution, budget constraint $B_i$ and PPC constraint $C_i$.

Our problem in (\ref{eqn:primal}) is a combination of the above global objective and private objectives. However, due to the coupled constraints and partial control of multiple ADs, obtaining an optimal bidding form directly for MACG seems intractable. We heuristically construct our optimal bidding formula as:
\begin{equation}
    b_{ij} = \lambda_0 \alpha_j \cdot CVR_{ij} \cdot IP_{ij} + \lambda_1 \beta_i \cdot \frac{g_k(i,j)}{CTR_{ij}} + \lambda_2 \gamma_i C_i, \forall i \in \mathcal{I},
    \label{base}
\end{equation}
where $(\lambda_0, \lambda_1, \lambda_2)$ are undetermined weight parameters to be learned. We argue that the above formula forms a superset for the Pareto optimal solution set of problem (\ref{eqn:primal}).

\begin{figure*}[htbp]
\centering 
\includegraphics[height=6.0cm]{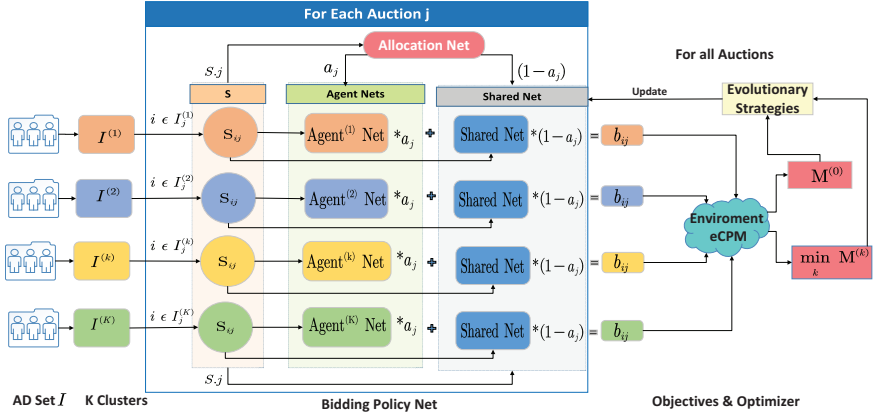} 
\caption{The architecture of MACG. $|\mathcal{I}|$ ADs are firstly clustered by their objectives: $\{\mathcal{I}^{(k)}\},\forall k$. Under each auction $j$, each AD $i \in \mathcal{I}_j^{(k)}$ obtains an initial bid ${(b_{AD})}_{ij}$ from $Agent^{(k)} \ Net$ based on features vector $\mathbf{S_{ij}}$. The bid is further modified by the output ${{(b_{0})}_{ij}}$ of a shared net which aims to achieve the global objective by taking into account the summarized features $\mathbf{S_{.j}}$. This modification is achieved by an $Allocation \ Net$ by using $\mathbf{S_{.j}}$ to assess real-time interpolation weights for $Agent^{(k)} \ Net$ and $shared \ Net$. We use ES to update the policy net based on objective scores $M^{(0)}$ and $\{M^{(k)}\}$ accumulated in an episode.} 
\label{Fig.main2} 
\end{figure*}

\section{METHODOLOGY}
The overall architecture of MACG is shown in Fig.~\ref{Fig.main2}. The ADs are clustered by their objectives. ADs with the same objective share the same agent model (Agent Net in the policy net). Agent sharing helps save parameter space and avoid overfitting (ADs in the long-tail show little training information). The policy net is designed according to the optimal bidding formula in Eq.(\ref{base}) and will be described in Section~\ref{section:policynet}. Section~\ref{sec:objective} discusses objective scores and model optimization.

\subsection{Bidding Policy Net}\label{section:policynet}
As aforementioned, the optimal bidding formula in Eq.(\ref{base}) is generally a combination of two parts, accounting for selfish and global objectives respectively. Motivated by this, under each auction $j \in \mathcal{J}$ each AD $i \in \mathcal{I}^{(k)}_j$ obtains its real-time bid ${b_{ij}}$ from the outputs of $Agent^{(k)} \ Net$ (selfish bid) and $Shared \ Net$ (cooperative bid), which are properly weighted by the output of $Allocated \ Net$. $\mathbf{S_{ij}}$ is the input feature vector of AD $i$ and $\mathbf{S_{.j}}$ represents the summarized features for $\mathcal{I}_j$, i.e., mean and variance of each feature in $\mathbf{S_{ij}}$ for ADs in $\mathcal{I}_j$. The detail features are described in the appendix. Next, we will elaborate the design of these sub-models.

\subsubsection{Agent Net}
The agent-specific network $Agent^{(k)} \ Net$ (abbr. $AG^{(k)}$) for each objective $k$ takes $\mathbf{S_{ij}}$ as input. Firstly, its benchmark bid refers to the term related to selfish objective in Eq.\ref{base}: $\frac{g_{k}(i,j)}{CTR_{ij}}$. In order to fit the coefficients in Eq.(\ref{base}) and expand the solution space, we use a neural network $y^{(k)}(\mathbf{S_{ij}})$ to generate a personalized correction factor for AD $i$ based on $\mathbf{S_{ij}}$. The finally formula of $AG^{(k)}$ to generate selfish bid ${(b_{AD})}_{ij}$ for an AD $i \in \mathcal{I}^{(k)}_j$ is: 
\begin{center}
\begin{equation}
\begin{split}
{(b_{AD})}_{ij} &= AG^{(k)}(i,j,\mathbf{S_{ij}}) = \frac{g_{k}(i,j)}{CTR_{ij}}*y^{(k)}(\mathbf{S_{ij}}) 
\end{split}
\end{equation}
\end{center}
Since $y^{(k)}(\mathbf{S_{ij}})$ is a multiplicative factor, we use the sigmoid function and transformation to normalize the output of the neural network to keep $y^{(k)}(\mathbf{S_{ij}})$ in $[1-Range,1+Range]$, where $Range$ is a hyperparameter.


\subsubsection{Shared Net}
In order to optimize the overall GMV of ADs, the $Shared \ Net$ (abbr. $SH$) takes AD-specific features $\mathbf{S_{ij}}$ (only $CVR_{ij}$ and $IP_{ij}$) and the summarized features $\mathbf{S_{.j}}$ as input, and generates a bid for an AD $i$ from the global objective's perspective. It refers to the term related to global objective (i.e., $CVR_{ij}*IP_{ij}$) in Eq.(\ref{base}) as benchmark. Furthermore, to portray the overall bidding environment, we introduce the average value $\overline{tk}_{.j}$ of $\{tk_{i}\}_{\forall i \in \mathcal{I}_j}$, where $tk_{i}$ is the reciprocal of AD $i$'s ROI and is calculated based on historical purchase transactions of AD $i$ as accumulated GMV divided by accumulated cost. Similar to the Agent net, we also use a neural network $y^{(0)}(\mathbf{S_{.j}})$ as a correction factor for calculating the bid from global perspective:
\begin{center}
\begin{equation}
\begin{split}
{(b_{0})}_{ij} &= SH(i,j,\mathbf{S_{ij}},\mathbf{S_{.j}}) = CVR_{ij}*IP_{ij}*\overline{tk}_{.j}*y^{(0)}(\mathbf{S_{.j}}) 
\end{split}
\end{equation}
\end{center}
where $y^{(0)}(\mathbf{S_{.j}})$ is again forced to be in $[1-Range,1+Range]$. The decomposition of $\alpha_j$ of (\ref{base}) into $\overline{tk}_{.j}$ and $y^{(0)}(\mathbf{S_{.j}})$ here can be seen as a normalization process to reduce the search space of $\alpha_j$. Next, we show the connection between ${(b_{0})}_{ij}$ and optimizing global objective. If we assume that AD ${i}$ only bids based on ${(b_{0})}_{ij}$, we can obtain the following ranking score in auction $j$ according to $eCPM$:
\begin{equation}
\begin{split}
eCPM_{ij} &= CTR_{ij} * CVR_{ij}*IP_{ij}*\overline{tk}_{.j}*y^{(0)}(\mathbf{S_{.j}}) \\ &= GMV_{ij}* \overline{tk}_{.j}*y^{(0)}(\mathbf{S_{.j}}) 
\end{split}
\end{equation}
Note that $\overline{tk}_{.j}*y^{(0)}(\mathbf{S_{.j}})$ is the same for all the involved ADs, which means we are intrinsically ranking ADs by $GMV_{ij}$. Therefore, $Shared \ Net$ naturally favor ADs that lead to high GMV.

\subsubsection{Allocation Network}
The intuitions behind the Allocation Network are: (1) It could make the exploration of the solution space more flexible. (2) When trained to optimize the multi-objectives, it could intelligently assess the relative importance between ${(b_{AD})}_{ij}$ and ${(b_{0})}_{ij}$. For instance, when the $GMV_{ij}$'s for ADs in $\mathcal{I}_j$ are similar (i.e., who wins the auction does not affect the overall GMV), it would emphasize ${(b_{AD})}_{ij}$ for these ADs so as to optimize their respective objectives. Formally, let ${AL}$ denote the function of the network and $a_j = {AL}(\mathbf{S_{.j}}) \in (0,1)$ be the output real-time weight to assess the importance of ${(b_{AD})}_{ij}$. The final bid $b_{ij}$ generated by the Policy Net for each AD $i \in \mathcal{I}_j$ is computed as follows:
\begin{center}
\begin{equation}
b_{ij} = a_j*{(b_{AD})}_{ij} + (1-a_j)*{{(b_{0})}_{ij}}
\label{new}
\end{equation}
\end{center}
Note in the optimal bidding formula of Eq.(\ref{base}) the third term $\gamma_i*C_i$ only refers to the constant $C_i$ in the constraints. Hence, it can be easily accounted for by the generated factors of the policy net.


\subsection{Objective Scores and Optimization}\label{sec:objective}

\subsubsection{Objective Scores}
In this paper, we consider three popular objectives\footnote{They are popular combinations of objectives and constraints on Taobao. This makes the scopes of AD constraints slightly different from those of Eq.~(\ref{eqn:primal}).} of ADs: optimizing click volume under the constraint of PPC (\textbf{Objective 1}), optimizing GMV under the constraint of cost (\textbf{Objective 2}) and optimizing volume of adding to shopping cart (CART) under the constraint of cost (\textbf{Objective 3}). The objective scores of these objectives are calculated by accumulation in one episode, i.e. $\mathcal{J}$. For clarity, we first define an accumulating function: 
\begin{equation}
    a(\{e_{ij}\}, k, V) = \sum\limits_{j \in \mathcal{J}}\sum\limits_{i \in \mathcal{I}^{(k)}_j}{e_{ij}*V_{ij}}
\end{equation}
where $\{e_{ij}\}$ represents the RTB results of our policy net, $V$ denotes the target value to accumulate and $k$ means we accumulate only among ADs with objective $k$. For example, $a(\{e_{ij}\}, 1, CTR)$ means we accumulate the CTR values of winning ADs with objective 1, corresponding to the predicted click volume of these ADs in an episode\footnote{Note that since training can only be done on offline data by simulation, the objective scores are calculated by predicted results.}; $a(\{e_{ij}\}, 2, w)$ means the accumulated cost of ADs with objective 2.

The design ideas for objective scores are as follows: (1) They should take the corresponding constraints into consideration, e.g., as regularization penalties. (2) Since ADs with the same objective share the same agent model, it is more reasonable to take them as a whole. (3) to facilitate optimization we should keep different objectives/constraints in a similar scale. Hence, we define objective scores using relative ratio values between two bidding policies: the optimal bidding policy to be learned vs. the benchmark bidding policy. In this work, we take OCPC \cite{zhu2017optimized} as the benchmark bidding policy. Let $\{e_{ij}^{'}\}$ represent the RTB results of the OCPC policy. The objective score for objective 1 is defined as:
\begin{equation}
    M^{(1)} = \frac{a(\{e_{ij}\}, 1, CTR)}{a(\{e_{ij}^{'}\}, 1, CTR)}  - \frac{a(\{e_{ij}\}, 1, w) / a(\{e_{ij}\}, 1, CTR)}{a(\{e_{ij}^{'}\}, 1, w) / a(\{e_{ij}^{'}\}, 1, CTR)}
    \label{click}
\end{equation}
Note that Eq.~(\ref{click}) takes the form of "objective $-$ constraint", where the first and second terms are ratios of aggregated click volume and PPC, respectively, between the two polices. The score definitions for the other objectives, Eqs.~(\ref{GMV_AD}) and~(\ref{CART}), are similar to Eq.~(\ref{click}):
\begin{equation}
    M^{(2)} = \frac{a(\{e_{ij}\}, 2, GMV)}{a(\{e_{ij}^{'}\}, 2, GMV)}  - \frac{a(\{e_{ij}\}, 2, w)}{a(\{e_{ij}^{'}\}, 2, w)}
    \label{GMV_AD}
\end{equation}

\begin{equation}
    M^{(3)} = \frac{a(\{e_{ij}\}, 3, CART)}{a(\{e_{ij}^{'}\}, 3, CART)}  - \frac{a(\{e_{ij}\}, 3, w)}{a(\{e_{ij}^{'}\}, 3, w)}
    \label{CART}
\end{equation}

As for the global objective, we maximize global GMV under the constraint to keep the platform revenue (i.e., cost of ADs). The corresponding objective scores is defined as follows:
\begin{equation}
    M^{(0)} = \frac{a(\{e_{ij}\}, 0, GMV)}{a(\{e_{ij}^{'}\}, 0, GMV)} -1  - |\frac{a(\{e_{ij}\}, 0, w)}{a(\{e_{ij}^{'}\}, 0, w)} -1 |
    \label{GMV}
\end{equation}
Note that $k=0$ in the accumulating function means we aggregate w.r.t ADs in $\mathcal{I}^{(0)}$ where we define $\mathcal{I}^{(0)}= \mathcal{I} \cup \overline{\mathcal{I}}$, i.e., the whole AD set containing self-bidding ADs. The constraint is introduced in a different way. When the platform revenue drops below that of the OCPC policy, the ``$-$'' between the two terms turns into ``$+$'' due to the absolute value operation, which means we should increase the revenue to avoid collusion among ADs. If it goes beyond OCPC, we also punish it since advertisers would not be happy to achieve higher profit with higher cost. In order to be consistent with the $M^{(1)}$, $M^{(2)}$ and $M^{(3)}$ ratio, we subtract 1 from the target.


\subsubsection{Model Optimization}
In order to reduce the search space of multi-objective optimization and keep optimization flexible for different numbers of ADs' objectives, we propose to optimize ADs' multiple objectives by focusing on the minimum one in each iteration. The detailed mathematical definition is as follows:
\begin{equation}
M_{AD} = min(M^{(1)}, M^{(2)}, M^{(3)})
\label{M_AD}
\end{equation}
Based on this optimization strategy, we transform the problem into optimizing the minimum score of ADs' objectives together with the score for global objective in each iteration. In order to achieve better efficiency for large-scale industrialization, we further combine them by introducing a hyperparameter $\lambda_M$:
\begin{equation}
M_{all} = M^{(0)} + \lambda_M*M_{AD}
\label{M_all}
\end{equation}
$M_{all}$ will be used as the evaluation score for seeking optimal parameters in ES.

Let $\mathbf{\theta}$ denote the vector containing all the parameters of the bidding policy network. We first initialize a set of $H$ parameter vectors $\{\theta_{ph}\}_{h=1}^{H}$ from the standard normal distribution for the first iteration $p=1$, and simulate RTB by an offline simulator for each $\theta_{1h}$. In the appendix, we will elaborate the offline simulator in detail. We then calculate the $M_{all}$ scores for these parameter vectors, and select the $W$ parameter vectors with the highest $M_{all}$ scores (for $p \geq 2$, we first filter these parameter vectors by retaining only those with both $M^{(0)}$ and $M_{AD}$ scores no less than the best results in the last iteration. In this way, we can guarantee the best performing $\theta$ in each iteration does not have degenerated objective scores, and converges to a Pareto optimal solution with $p$ increasing.). For the selected parameter vectors, we perform random perturbation (i.e. $\epsilon_{h} \sim N(0,I)$ where $\epsilon_{h}$ is a perturbation parameter) to generate a new population $\{\theta_{(p+1)h}\}_{h=1}^{H}$ of parameter vectors for the next iteration. The above process is repeated until the $M_{all}$ score converges and finally we select the top parameter vector as the optimal model. With such a parameter exploration scheme, the training would converge faster compared to the previous RL-based methods based on action space exploration, since ES is less affected by the complex bidding environment. In practice, we implement the training of MACG in a distributed computation environment. The pseudo-code is shown in Algorithm 1 in the appendix.



\begin{table*}
\center \caption{Results for the offline simulator across different days. The percentage values denote the improvement ratios of the methods versus OCPC.}\label{offline_comparison}
\begin{tabular}{|c|c|c|c|c|c|c|c|c|}
\hline 
 & \multicolumn{4}{|c|}{August 9, 2020} & \multicolumn{4}{|c|}{July 16,2020}  \\
 \hline
  & $M^{(0)}$ & $M^{{(1)}}$ & $M^{{(2)}}$ & $M^{{(3)}}$ & $M^{(0)}$ & $M^{{(1)}}$ & $M^{{(2)}}$ & $M^{{(3)}}$\\
  \hline
 OCPC &100\% &100\% &100\% & 100\% &100\% &100\% &100\% & 100\% \\
 \hline
 MKB     &86.7\% &87.6\% &88.1\% & 87.4\% &87.1\% &86.6\% &85.8\% & 86.3\% \\
 \hline
 M-RMDP &104.5\% &  94.5\% &104.3\% & 98.7\% &103.4\% & 93.6\% &103.3\% & 97.5\% \\
 \hline
 DCMAB &106.7\% & 96.6\% &106.4\% & 99.4\% &105.9\% & 95.6\% &105.7\% & 98.7\% \\
 \hline
  MACG &$\mathbf{111.6\%}$ &$\mathbf{106.2\%}$ &$\mathbf{107.2\%}$& $\mathbf{112.9\%}$ &$\mathbf{110.4\%}$ &$\mathbf{107.5\%}$ &$\mathbf{107.1\%}$& $\mathbf{113.2\%}$\\
 \hline
\end{tabular}
\end{table*}

\section{EXPERIMENTS}
In this section, we evaluate the proposed bid optimization algorithm in MACG by conducting a comprehensive suite of experiments on a large-scale offline dataset, and via online A/B tests in Taobao.
\subsection{Dataset}
We collected a real-world dataset from Taobao sponsored search platform for offline evaluation. The dataset contains auctions data and ADs' features, of 4 days (July 15th, 2020, July 16th, 2020, August 8th, 2020 and August 9th, 2020). For each day, we randomly sampled $10^9$ auctions with nearly $10^6$ ADs. We perform two parallel groups of experiments and treat the data of July 15th (August 8th) as the training set, and the corresponding next day, July 16th (August 9th) as the test set for the first (second) group. The percentages of ADs with the GMV, CART and click objectives are 52\%, 11\% and 38\%, respectively. 

\subsection{Compared Methods and Evaluation Setting}
Our MACG is compared with the following methods.

\noindent \textbf{Manual Keyword-level Bidding} (MKB) sets bids on each keyword and all the auctions corresponding to the keyword use the fixed bids.

\noindent \textbf{Optimized Cost Per Click} (OCPC) \cite{zhu2017optimized} ranks the multiple ADs with respect to the optimization objective. Then the ranked AD list is examined sequentially to check if the required bids lie in the feasible bid ranges. OCPC is a single-auction solution with only bid limits constraint. In our experiments, we treat OCPC as the fundamental baseline with GMV as the optimization objective.

\noindent \textbf{Massive-agent RL with Robust MDP} (M-RMDP) \cite{zhao2018deep} is an RL-based method which considers each AD as an agent and trains a multi-agent deep Q-learning network \cite{gu2016continuous} to bid. In order to apply it to our scenario with millions of ADs, we divide ADs into multiple clusters according to different objectives, and then assign an agent to an AD cluster.

\noindent \textbf{Distributed Coordinated Multi-Agent Bidding} (DCMAB) \cite{lowe2017multi} firstly groups ADs and constructs an agent for each cluster. Then all agents’ bidding actions are fed into the critic function Q. Finally, a distributed coordinated multi-agent deep deterministic policy gradient technique is utilized to update the actor network.

M-RMDP and DCMAB were proposed for the single objective case. As aforementioned, RL-based methods suffer from poor convergence due to very complex environment. And the situation becomes much worse when handling multi-objective optimization. Thereby, we set their objective as maximizing GMV under budget constraints and split all-day auction data into 24 time slices, which is consistent with their experiment settings. For fair comparison, we set an agent for each AD cluster ($\mathcal{I}^{(1)}$, $\mathcal{I}^{(2)}$, $\mathcal{I}^{(3)}$). 

We evaluate the performance from two aspects, i.e., ADs' objectives and the global objective. For ADs' objectives including \emph{the click volume under PPC constraint}, \emph{GMV under cost constraint} and \emph{volume of CART under cost constraint}, we assess them on the corresponding AD clusters $\mathcal{I}^{(1)}$, $\mathcal{I}^{(2)}$, $\mathcal{I}^{(3)}$ by scores $M^{(1)}$, $M^{(2)}$ and $M^{(3)}$ defined in Section \ref{sec:objective}, respectively. $M^{(0)}$ calculated on the whole set of ADs is utilized to evaluate the global objective. We report the improvement ratios of all the methods compared with OCPC.


\begin{table*}[!t]
\center \caption{Results of offline ablation experiments across different days. The percentage values denote the improvement ratios of the methods versus OCPC.}\label{Ablation}
\begin{tabular}{|c|c|c|c|c|c|c|c|c|}
\hline 
 & \multicolumn{4}{|c|}{August 9, 2020} & \multicolumn{4}{|c|}{July 16,2020}  \\
 \hline
  & $M^{(0)}$ & $M^{{(1)}}$ & $M^{{(2)}}$ & $M^{{(3)}}$ & $M^{(0)}$ & $M^{{(1)}}$ & $M^{{(2)}}$ & $M^{{(3)}}$\\
  \hline
 OCPC &100\% &100\% &100\% & 100\% &100\% &100\% &100\% & 100\% \\
 \hline
 MACG-g         &105.3\% &104.9\% &106.4\% & 107.8\% &104.2\% &105.3\% &104.5\% & 108.1\%\\
 \hline
 MACG-l         &$\mathbf{112.4}\%$ & 101.3\% & $\mathbf{109.9\%}$ & 103.2\% & $\mathbf{114.3\%}$ &102.1\% & $\mathbf{110.7\%}$ & 102.6\%\\
 \hline
 MACG-a         &110.4\% &105.6\% &106.3\% & 111.7\% &109.9\% &106.1\% &106.7\% & 112.8\%\\
 \hline
  MACG & 111.6\% &$\mathbf{106.2\%}$ & 107.2\% & $\mathbf{112.9\%}$ & 110.4\% &$\mathbf{107.5\%}$ & 107.1\% & $\mathbf{113.2\%}$\\
 \hline
\end{tabular}
\end{table*}

\begin{figure}[t]
\subfigure[August 8th, 2020]{\label{w}\includegraphics[height=3.3cm]{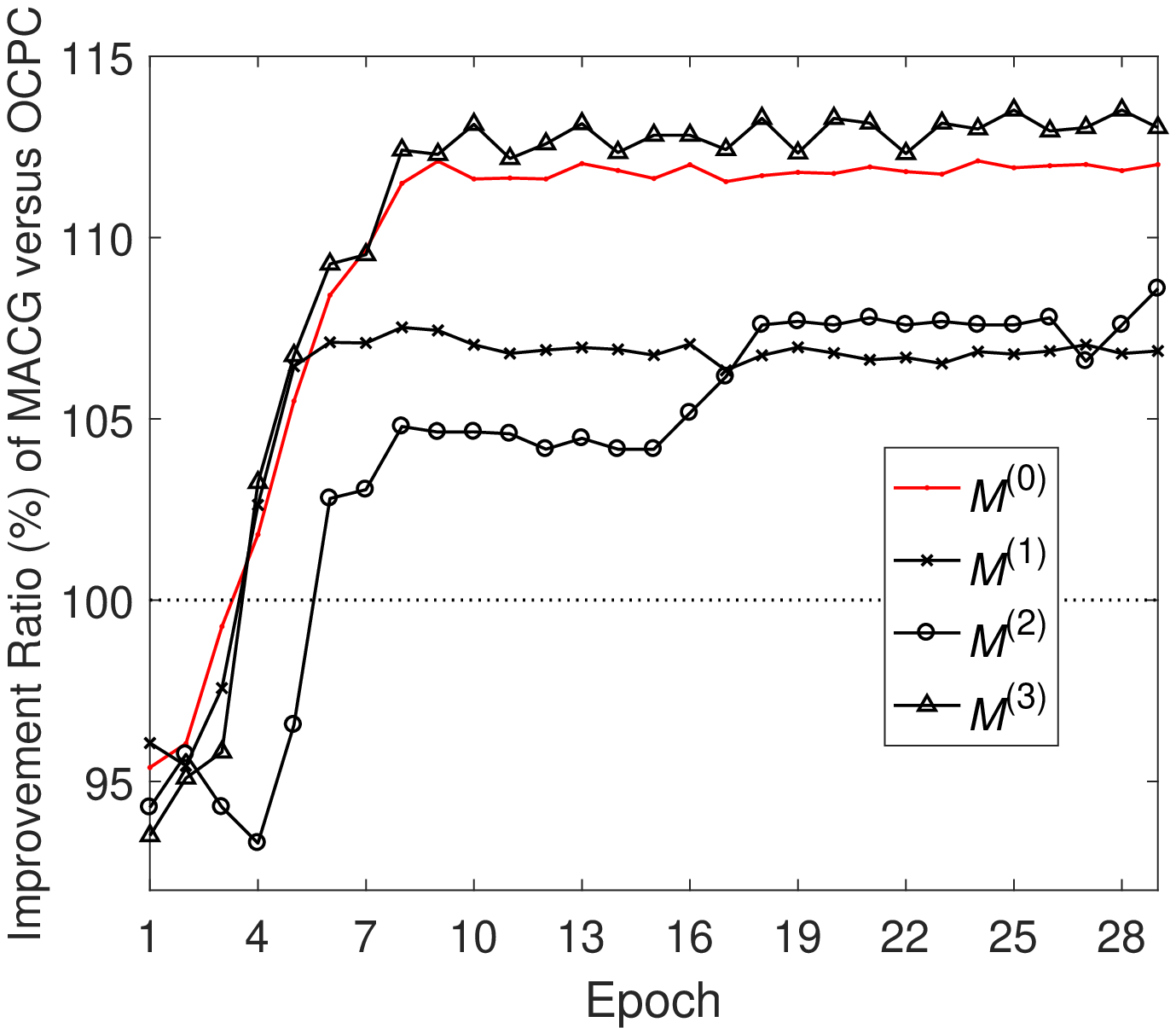}}
\subfigure[July 15th, 2020]{\label{q}\includegraphics[height=3.3cm]{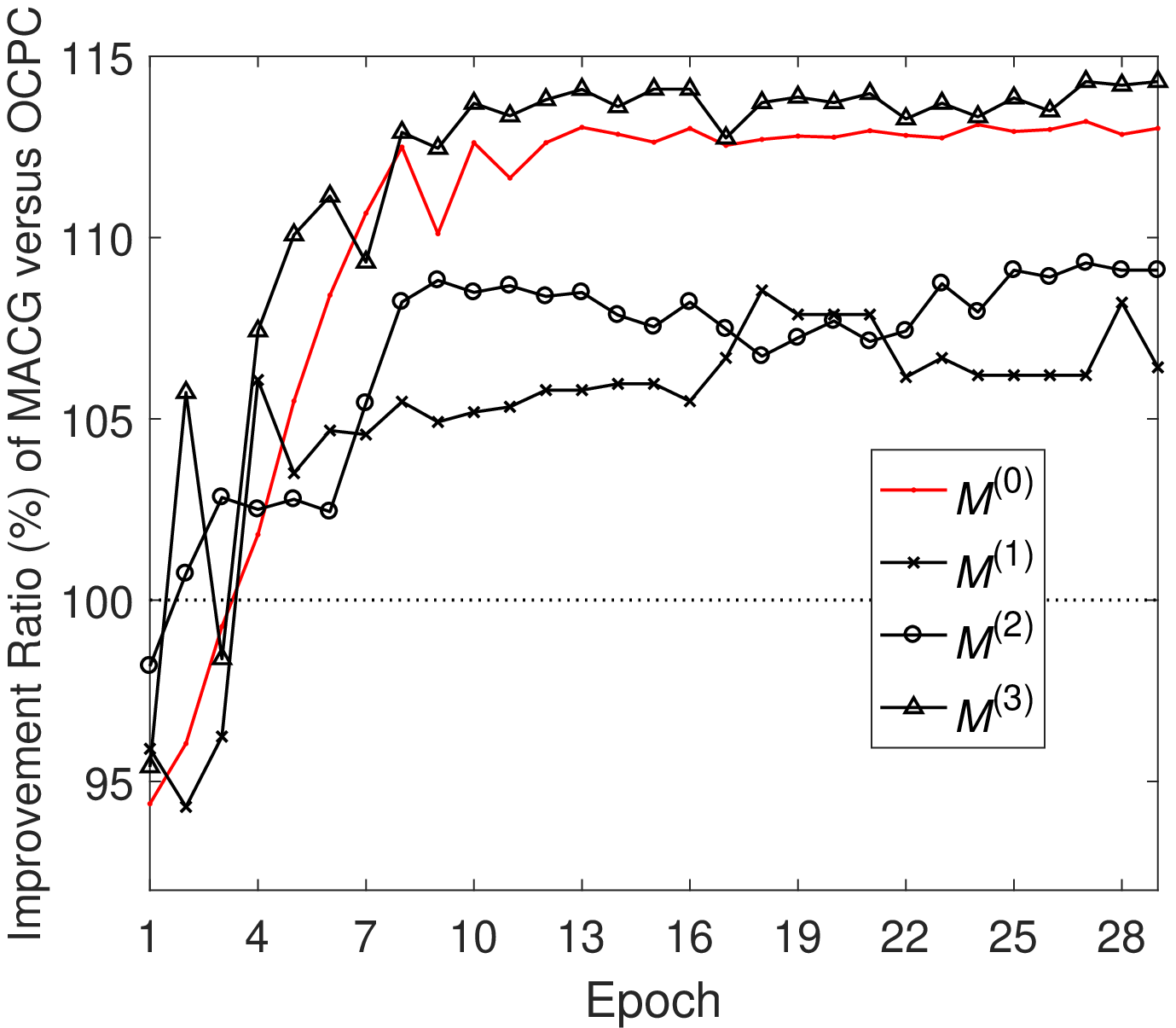}}
\caption{Training convergence on August 8th, 2020 and July 15th, 2020.} \label{convergence}
\end{figure}

\subsection{OFFLINE EVALUATION}



In the offline experiments, for all auctions $\mathcal{J}$ in the test set, we first obtain the winning ADs of different methods through offline simulator. Then we combine the historical data, such as $CTR$ and $CVR$ to calculate the estimated objective scores. 

\subsubsection{Model Comparison}
Tab. \ref{offline_comparison} shows the experimental results of different methods. First, MKB performs the worst. This reveals the importance of auction-level real time bidding. Second, RL based methods (M-RMDP, DCMAB) outperform OCPC on the $M^{(0)}$ and $M^{(2)}$ scores. This is intuitive since OCPC learns the optimal bidding strategy for each AD separately in each auction, and therefore fails to consider bidding interactions in the complex auction environment. On the contrary, M-RMDP and DCMAB use multi-agent DQN and multi-agent DDPG to characterize the complex bidding interactions in this environment. Third, the $M^{(1)}$ and $M^{(3)}$ scores of M-RMDP and DCMAB exhibit large differences compared with their $M^{(0)}$ and $M^{(2)}$ scores. The reason might be that $M^{(0)}$ and $M^{(2)}$ correspond to the global and ADs' GMV, which is also the optimization objective of M-RMDP and DCMAB. However, $M^{(1)}$ and $M^{(3)}$ scores evaluate ADs' other objectives. The results show that the single objective methods are difficult to satisfy ADs' personalized objectives. Finally, MACG significantly outperforms all the compared methods on all the objectives. This confirms that the proposed multi-objective cooperative bidding strategy improves both single advertiser's objective and global profit.

\begin{figure*}[t]
\centering 
\includegraphics[width=5in]{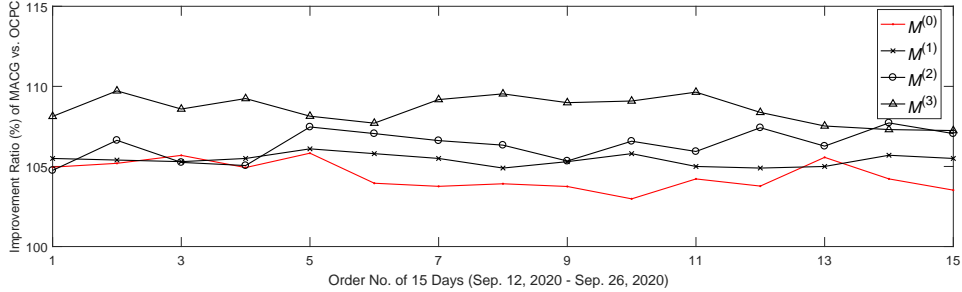} 
\caption{Online experiment results on continuous 15 days (Sep. 12th, 2020 - Sep. 26th, 2020).} 
\label{online} 
\end{figure*}

\subsubsection{Ablation Experiments}
Next we conduct ablation experiments to verify the effectiveness of different parts of MACG, including the sub-networks for the global objective and ADs' objectives, as well as the allocation network: 1) \textbf{MACG-g (MACG without shared net)}. To prove the global objective can achieve better global profit with the shared network of MACG, we remove the shared network and the allocation network to construct MACG-g. The allocation network is designed to allocate the weights of shared network and agent networks. When the shared network or (agent networks) is removed, the allocation network will naturally be removed. 2) \textbf{MACG-l (MACG without agent nets)}. To verify the effectiveness of our multi-objective cooperative strategy, we remove the agent networks and the allocation network to construct another variant, i.e., MACG-l. This corresponds to sorting smart-bidding ADs always according to GMV. Due to the existence of self-bidding ADs, we could still optimize the effect of smart-bidding ADs by optimizing the shared network. 3) \textbf{MACG-a (MACG without the allocation net)}. In MACG-a, the output of the allocation network is replaced by a single parameter $a$. 

Tab. \ref{Ablation} shows the results. There are two important observations: 1) MACG outperforms MACG-g and MACG-a on all evaluation scores. The global score $M^{(0)}$ of MACG-g drops sharply. This indicates without shared net, ADs more easily suffer ineffective competition. In addition, the evaluation scores for ADs' objectives also decrease. The reason is that compared with the multi-AD cooperation in MACG, the competition mode in MACG-g causes a smaller solution space. This influences model optimization. MACG-a uses static weight parameters $a$, which cannot fit the real-time auction environment. 2) For MACG-l, only the global GMV objective is optimized, resulting in the improvement in GMV related scores, $M^{(0)}$ and $M^{(2)}$, but the other scores are influenced seriously. 



\subsubsection{Convergence Analysis}
The multi-objective optimization problem in MACG is solved by evolution strategies. Here we analyse its empirical convergence properties. Fig.~\ref{convergence} shows the global objective and ADs' multi-objective scores against the number of iterations. We find that at the beginning the objective scores increase rapidly and then become stable in about 10 iterations. Experimental results verify that MACG has a strong learning ability in exploring the optimal solutions. The reason could be that the search space of MACG is small (the parameter number of MACG is less than 200, which is elaborated in the Appendix). Therefore, MACG can efficiently explore the optimal solutions via parameter space exploration. 

\subsection{ONLINE EVALUATION}
We now test MACG in the real auction environment of the Alibaba search auction platform. Nearly $10^7$ ADs participate in the online real-time bidding, with $10^{11}$ auctions requests on an ordinary day. Considering the volatility of online traffic, we conduct the online experiments from Sep. 12th, 2020 to Sep. 26th, 2020 (15 days in total). For model training, we use the records of the previous day to update the parameters in day level. We use a standard A/B testing configuration and still consider OCPC as the benchmark bucket. We randomly sample 20\% online traffics as the experimental bucket and another 20\% online traffics as the benchmark bucket.

\subsubsection{Result Analysis}
Fig.~\ref{online} shows the online experiment results and we can get the following observations: 1) MACG still significantly outperforms the baseline method. For instance, MACG achieves on average 4.42\% improvement of $M^{{(0)}}$. 2) MACG maintains a relatively stable improvement ratio in 15 days (The standard deviations of $M^{{(0)}}$, $M^{{(1)}}$, $M^{{(2)}}$ and $M^{{(3)}}$ in this period are 0.75\%, 0.34\%, 0.91\% and 0.84\%, respectively). This proves the effectiveness and robustness of our model in the online environment.


\subsubsection{Experimental Difference Analysis}
Note that there exist performance differences for MACG between online and offline experiments. For example, the online $M^{(0)}$ scores are about 5\% lower than offline scores. The reasons might be: 1) The objective scores in offline environment are calculated based on predicted values, and there could be a gap between the predicted values and the real results. For instance, the predicted GMV score depends on predicted CVR ($GMV = CTR * CVR * IP$). However, since the transaction behavior is extremely sparse, the estimation error in CVR prediction is inevitable. 2) In the offline data collection, if an AD's budget is used up, the AD will quit and the follow-up auctions do not involve this AD. The optimization of the model is somehow affected by this phenomenon in the offline simulator, while this phenomenon does not exist in the online setting. Such a discrepancy could also lead to degenerated performance.

\section{CONCLUSIONS}
In this paper, we propose a novel multi-AD cooperative games formulation for bid optimization in e-commercial sponsored search. MACG considers not only multi-objectives of smart-bidding advertisers but also the benefits of self-bidding advertisers and the platform. Furthermore, we theoretically analyze the optimal functional bidding formula under MACG. A multi-agent policy net is designed based on the optimal bidding form to solve MACG, encoding both selfish and global objectives. In order to achieve timely training of the model for large-scale industrial environments, we heuristically design an efficient multi-agent ES to update network parameters. For empirically evaluation, we deploy MACG in a distributed computation environment in Taobao's search auction platform and offline evaluation and standard online A/B tests prove the superiority of MACG compared to state-of-art methods. In the future, we will explore more efficient multi-objective optimization algorithms to pursue optimal solutions.
\section{ACKNOWLEDGMENTS}
This research was supported by the National Natural Science Foundation of China (Grant Nos. 61936006, 61876144, 61876145, 62073255), the Key Research and Development Program of Shaanxi (Program No.2020ZDLGY04-07), and Innovation Capability Support Program of Shaanxi (Program No. 2021TD-05). This work was also supported by Alibaba Group through Alibaba Research Intern Program. We hereby give specical thanks to Alibaba Group for their contribution to this paper.
\bibliographystyle{ACM-Reference-Format}
\bibliography{MAES}


\begin{thebibliography}{28}


\ifx \showCODEN    \undefined \def \showCODEN     #1{\unskip}     \fi
\ifx \showDOI      \undefined \def \showDOI       #1{#1}\fi
\ifx \showISBNx    \undefined \def \showISBNx     #1{\unskip}     \fi
\ifx \showISBNxiii \undefined \def \showISBNxiii  #1{\unskip}     \fi
\ifx \showISSN     \undefined \def \showISSN      #1{\unskip}     \fi
\ifx \showLCCN     \undefined \def \showLCCN      #1{\unskip}     \fi
\ifx \shownote     \undefined \def \shownote      #1{#1}          \fi
\ifx \showarticletitle \undefined \def \showarticletitle #1{#1}   \fi
\ifx \showURL      \undefined \def \showURL       {\relax}        \fi
\providecommand\bibfield[2]{#2}
\providecommand\bibinfo[2]{#2}
\providecommand\natexlab[1]{#1}
\providecommand\showeprint[2][]{arXiv:#2}

\bibitem[\protect\citeauthoryear{Borgs, Chayes, Immorlica, Jain, Etesami, and
  Mahdian}{Borgs et~al\mbox{.}}{2007}]%
        {borgs2007dynamics}
\bibfield{author}{\bibinfo{person}{Christian Borgs}, \bibinfo{person}{Jennifer
  Chayes}, \bibinfo{person}{Nicole Immorlica}, \bibinfo{person}{Kamal Jain},
  \bibinfo{person}{Omid Etesami}, {and} \bibinfo{person}{Mohammad Mahdian}.}
  \bibinfo{year}{2007}\natexlab{}.
\newblock \showarticletitle{Dynamics of bid optimization in online
  advertisement auctions}. In \bibinfo{booktitle}{\emph{WWW}}.
  \bibinfo{pages}{531--540}.
\newblock


\bibitem[\protect\citeauthoryear{Cai, Ren, Zhang, Malialis, Wang, Yu, and
  Guo}{Cai et~al\mbox{.}}{2017}]%
        {cai2017real}
\bibfield{author}{\bibinfo{person}{Han Cai}, \bibinfo{person}{Kan Ren},
  \bibinfo{person}{Weinan Zhang}, \bibinfo{person}{Kleanthis Malialis},
  \bibinfo{person}{Jun Wang}, \bibinfo{person}{Yong Yu}, {and}
  \bibinfo{person}{Defeng Guo}.} \bibinfo{year}{2017}\natexlab{}.
\newblock \showarticletitle{Real-time bidding by reinforcement learning in
  display advertising}. In \bibinfo{booktitle}{\emph{WSDM}}.
  \bibinfo{pages}{661--670}.
\newblock


\bibitem[\protect\citeauthoryear{Cavallo, Sviridenko, and Wilkens}{Cavallo
  et~al\mbox{.}}{2018}]%
        {cavallo2018matching}
\bibfield{author}{\bibinfo{person}{Ruggiero Cavallo}, \bibinfo{person}{Maxim
  Sviridenko}, {and} \bibinfo{person}{Christopher~A Wilkens}.}
  \bibinfo{year}{2018}\natexlab{}.
\newblock \showarticletitle{Matching auctions for search and native ads}. In
  \bibinfo{booktitle}{\emph{ACM EC}}. \bibinfo{pages}{663--680}.
\newblock


\bibitem[\protect\citeauthoryear{Deb, Pratap, Agarwal, and Meyarivan}{Deb
  et~al\mbox{.}}{2002}]%
        {deb2002fast}
\bibfield{author}{\bibinfo{person}{Kalyanmoy Deb}, \bibinfo{person}{Amrit
  Pratap}, \bibinfo{person}{Sameer Agarwal}, {and} \bibinfo{person}{TAMT
  Meyarivan}.} \bibinfo{year}{2002}\natexlab{}.
\newblock \showarticletitle{A fast and elitist multiobjective genetic
  algorithm: NSGA-II}.
\newblock \bibinfo{journal}{\emph{IEEE transactions on evolutionary
  computation}} \bibinfo{volume}{6}, \bibinfo{number}{2}
  (\bibinfo{year}{2002}), \bibinfo{pages}{182--197}.
\newblock


\bibitem[\protect\citeauthoryear{Evans}{Evans}{2009}]%
        {evans2009online}
\bibfield{author}{\bibinfo{person}{David~S Evans}.}
  \bibinfo{year}{2009}\natexlab{}.
\newblock \showarticletitle{The online advertising industry: Economics,
  evolution, and privacy}.
\newblock \bibinfo{journal}{\emph{Journal of economic perspectives}}
  \bibinfo{volume}{23}, \bibinfo{number}{3} (\bibinfo{year}{2009}),
  \bibinfo{pages}{37--60}.
\newblock


\bibitem[\protect\citeauthoryear{Geyik, Faleev, Shen, O'Donnell, and
  Kolay}{Geyik et~al\mbox{.}}{2016}]%
        {geyik2016joint}
\bibfield{author}{\bibinfo{person}{Sahin~Cem Geyik}, \bibinfo{person}{Sergey
  Faleev}, \bibinfo{person}{Jianqiang Shen}, \bibinfo{person}{Sean O'Donnell},
  {and} \bibinfo{person}{Santanu Kolay}.} \bibinfo{year}{2016}\natexlab{}.
\newblock \showarticletitle{Joint optimization of multiple performance metrics
  in online video advertising}. In \bibinfo{booktitle}{\emph{SIGKDD}}.
  \bibinfo{pages}{471--480}.
\newblock


\bibitem[\protect\citeauthoryear{Goldfarb and Tucker}{Goldfarb and
  Tucker}{2011}]%
        {goldfarb2011online}
\bibfield{author}{\bibinfo{person}{Avi Goldfarb} {and}
  \bibinfo{person}{Catherine Tucker}.} \bibinfo{year}{2011}\natexlab{}.
\newblock \showarticletitle{Online display advertising: Targeting and
  obtrusiveness}.
\newblock \bibinfo{journal}{\emph{Marketing Science}} \bibinfo{volume}{30},
  \bibinfo{number}{3} (\bibinfo{year}{2011}), \bibinfo{pages}{389--404}.
\newblock


\bibitem[\protect\citeauthoryear{Gu, Lillicrap, Sutskever, and Levine}{Gu
  et~al\mbox{.}}{2016}]%
        {gu2016continuous}
\bibfield{author}{\bibinfo{person}{Shixiang Gu}, \bibinfo{person}{Timothy
  Lillicrap}, \bibinfo{person}{Ilya Sutskever}, {and} \bibinfo{person}{Sergey
  Levine}.} \bibinfo{year}{2016}\natexlab{}.
\newblock \showarticletitle{Continuous deep q-learning with model-based
  acceleration}. In \bibinfo{booktitle}{\emph{ICML}}.
  \bibinfo{pages}{2829--2838}.
\newblock


\bibitem[\protect\citeauthoryear{Hansen and Ostermeier}{Hansen and
  Ostermeier}{2001}]%
        {hansen2001completely}
\bibfield{author}{\bibinfo{person}{Nikolaus Hansen} {and}
  \bibinfo{person}{Andreas Ostermeier}.} \bibinfo{year}{2001}\natexlab{}.
\newblock \showarticletitle{Completely derandomized self-adaptation in
  evolution strategies}.
\newblock \bibinfo{journal}{\emph{Evolutionary computation}}
  \bibinfo{volume}{9}, \bibinfo{number}{2} (\bibinfo{year}{2001}),
  \bibinfo{pages}{159--195}.
\newblock


\bibitem[\protect\citeauthoryear{Huning}{Huning}{1976}]%
        {huning1976evolutionsstrategie}
\bibfield{author}{\bibinfo{person}{Alois Huning}.}
  \bibinfo{year}{1976}\natexlab{}.
\newblock \bibinfo{title}{Evolutionsstrategie. optimierung technischer systeme
  nach prinzipien der biologischen evolution}.
\newblock
\newblock


\bibitem[\protect\citeauthoryear{Iruthayarajan and Baskar}{Iruthayarajan and
  Baskar}{2010}]%
        {iruthayarajan2010covariance}
\bibfield{author}{\bibinfo{person}{M~Willjuice Iruthayarajan} {and}
  \bibinfo{person}{S Baskar}.} \bibinfo{year}{2010}\natexlab{}.
\newblock \showarticletitle{Covariance matrix adaptation evolution strategy
  based design of centralized PID controller}.
\newblock \bibinfo{journal}{\emph{Expert systems with Applications}}
  \bibinfo{volume}{37}, \bibinfo{number}{8} (\bibinfo{year}{2010}),
  \bibinfo{pages}{5775--5781}.
\newblock


\bibitem[\protect\citeauthoryear{Jin, Song, Li, Gai, Wang, and Zhang}{Jin
  et~al\mbox{.}}{2018a}]%
        {jin2018multi}
\bibfield{author}{\bibinfo{person}{Junqi Jin}, \bibinfo{person}{Chengru Song},
  \bibinfo{person}{Han Li}, \bibinfo{person}{Kun Gai}, \bibinfo{person}{Jun
  Wang}, {and} \bibinfo{person}{Weinan Zhang}.}
  \bibinfo{year}{2018}\natexlab{a}.
\newblock \showarticletitle{Real-time bidding with multi-agent reinforcement
  learning in display advertising}. In \bibinfo{booktitle}{\emph{CIKM}}.
  \bibinfo{pages}{2193--2201}.
\newblock


\bibitem[\protect\citeauthoryear{Jin, Song, Li, Gai, Wang, and Zhang}{Jin
  et~al\mbox{.}}{2018b}]%
        {jin2018real}
\bibfield{author}{\bibinfo{person}{Junqi Jin}, \bibinfo{person}{Chengru Song},
  \bibinfo{person}{Han Li}, \bibinfo{person}{Kun Gai}, \bibinfo{person}{Jun
  Wang}, {and} \bibinfo{person}{Weinan Zhang}.}
  \bibinfo{year}{2018}\natexlab{b}.
\newblock \showarticletitle{Real-time bidding with multi-agent reinforcement
  learning in display advertising}. In \bibinfo{booktitle}{\emph{CIKM}}.
  \bibinfo{pages}{2193--2201}.
\newblock


\bibitem[\protect\citeauthoryear{Kitts and Leblanc}{Kitts and Leblanc}{2004}]%
        {kitts2004optimal}
\bibfield{author}{\bibinfo{person}{Brendan Kitts} {and}
  \bibinfo{person}{Benjamin Leblanc}.} \bibinfo{year}{2004}\natexlab{}.
\newblock \showarticletitle{Optimal bidding on keyword auctions}.
\newblock \bibinfo{journal}{\emph{Electronic markets}} \bibinfo{volume}{14},
  \bibinfo{number}{3} (\bibinfo{year}{2004}), \bibinfo{pages}{186--201}.
\newblock


\bibitem[\protect\citeauthoryear{Lowe, Wu, Tamar, Harb, Abbeel, and
  Mordatch}{Lowe et~al\mbox{.}}{2017}]%
        {lowe2017multi}
\bibfield{author}{\bibinfo{person}{Ryan Lowe}, \bibinfo{person}{Yi~I Wu},
  \bibinfo{person}{Aviv Tamar}, \bibinfo{person}{Jean Harb},
  \bibinfo{person}{OpenAI~Pieter Abbeel}, {and} \bibinfo{person}{Igor
  Mordatch}.} \bibinfo{year}{2017}\natexlab{}.
\newblock \showarticletitle{Multi-agent actor-critic for mixed
  cooperative-competitive environments}. In
  \bibinfo{booktitle}{\emph{NeurIPS}}. \bibinfo{pages}{6379--6390}.
\newblock


\bibitem[\protect\citeauthoryear{Perlich, Dalessandro, Hook, Stitelman, Raeder,
  and Provost}{Perlich et~al\mbox{.}}{2012}]%
        {perlich2012bid}
\bibfield{author}{\bibinfo{person}{Claudia Perlich}, \bibinfo{person}{Brian
  Dalessandro}, \bibinfo{person}{Rod Hook}, \bibinfo{person}{Ori Stitelman},
  \bibinfo{person}{Troy Raeder}, {and} \bibinfo{person}{Foster Provost}.}
  \bibinfo{year}{2012}\natexlab{}.
\newblock \showarticletitle{Bid optimizing and inventory scoring in targeted
  online advertising}. In \bibinfo{booktitle}{\emph{SIGKDD}}.
  \bibinfo{pages}{804--812}.
\newblock


\bibitem[\protect\citeauthoryear{Ribeiro, Ziviani, Moura, Hata, Lacerda, and
  Veloso}{Ribeiro et~al\mbox{.}}{2014}]%
        {ribeiro2014multiobjective}
\bibfield{author}{\bibinfo{person}{Marco~Tulio Ribeiro}, \bibinfo{person}{Nivio
  Ziviani}, \bibinfo{person}{Edleno Silva~De Moura}, \bibinfo{person}{Itamar
  Hata}, \bibinfo{person}{Anisio Lacerda}, {and} \bibinfo{person}{Adriano
  Veloso}.} \bibinfo{year}{2014}\natexlab{}.
\newblock \showarticletitle{Multiobjective pareto-efficient approaches for
  recommender systems}.
\newblock \bibinfo{journal}{\emph{ACM Transactions on Intelligent Systems and
  Technology (TIST)}} \bibinfo{volume}{5}, \bibinfo{number}{4}
  (\bibinfo{year}{2014}), \bibinfo{pages}{1--20}.
\newblock


\bibitem[\protect\citeauthoryear{Sehnke, Osendorfer, R{\"u}ckstie{\ss}, Graves,
  Peters, and Schmidhuber}{Sehnke et~al\mbox{.}}{2010}]%
        {sehnke2010parameter}
\bibfield{author}{\bibinfo{person}{Frank Sehnke}, \bibinfo{person}{Christian
  Osendorfer}, \bibinfo{person}{Thomas R{\"u}ckstie{\ss}},
  \bibinfo{person}{Alex Graves}, \bibinfo{person}{Jan Peters}, {and}
  \bibinfo{person}{J{\"u}rgen Schmidhuber}.} \bibinfo{year}{2010}\natexlab{}.
\newblock \showarticletitle{Parameter-exploring policy gradients}.
\newblock \bibinfo{journal}{\emph{Neural Networks}} \bibinfo{volume}{23},
  \bibinfo{number}{4} (\bibinfo{year}{2010}), \bibinfo{pages}{551--559}.
\newblock


\bibitem[\protect\citeauthoryear{Sun, Wierstra, Schaul, and Schmidhuber}{Sun
  et~al\mbox{.}}{2009}]%
        {sun2009efficient}
\bibfield{author}{\bibinfo{person}{Yi Sun}, \bibinfo{person}{Daan Wierstra},
  \bibinfo{person}{Tom Schaul}, {and} \bibinfo{person}{Juergen Schmidhuber}.}
  \bibinfo{year}{2009}\natexlab{}.
\newblock \showarticletitle{Efficient natural evolution strategies}. In
  \bibinfo{booktitle}{\emph{GECCO}}. \bibinfo{pages}{539--546}.
\newblock


\bibitem[\protect\citeauthoryear{Wang, Wei, Yan, Chen, and Du}{Wang
  et~al\mbox{.}}{2012}]%
        {wang2012multi}
\bibfield{author}{\bibinfo{person}{Yilei Wang}, \bibinfo{person}{Bingzheng
  Wei}, \bibinfo{person}{Jun Yan}, \bibinfo{person}{Zheng Chen}, {and}
  \bibinfo{person}{Qiao Du}.} \bibinfo{year}{2012}\natexlab{}.
\newblock \showarticletitle{Multi-objective optimization for sponsored search}.
  In \bibinfo{booktitle}{\emph{ADKDD}}. \bibinfo{pages}{1--9}.
\newblock


\bibitem[\protect\citeauthoryear{Wilkens, Cavallo, and Niazadeh}{Wilkens
  et~al\mbox{.}}{2017}]%
        {wilkens2017gsp}
\bibfield{author}{\bibinfo{person}{Christopher~A Wilkens},
  \bibinfo{person}{Ruggiero Cavallo}, {and} \bibinfo{person}{Rad Niazadeh}.}
  \bibinfo{year}{2017}\natexlab{}.
\newblock \showarticletitle{GSP: the cinderella of mechanism design}. In
  \bibinfo{booktitle}{\emph{WWW}}. \bibinfo{pages}{25--32}.
\newblock


\bibitem[\protect\citeauthoryear{Yang, Li, Wang, Wu, Tan, Xu, and Gai}{Yang
  et~al\mbox{.}}{2019}]%
        {yang2019bid}
\bibfield{author}{\bibinfo{person}{Xun Yang}, \bibinfo{person}{Yasong Li},
  \bibinfo{person}{Hao Wang}, \bibinfo{person}{Di Wu}, \bibinfo{person}{Qing
  Tan}, \bibinfo{person}{Jian Xu}, {and} \bibinfo{person}{Kun Gai}.}
  \bibinfo{year}{2019}\natexlab{}.
\newblock \showarticletitle{Bid optimization by multivariable control in
  display advertising}. In \bibinfo{booktitle}{\emph{SIGKDD}}.
  \bibinfo{pages}{1966--1974}.
\newblock


\bibitem[\protect\citeauthoryear{Zhang, Yuan, and Wang}{Zhang
  et~al\mbox{.}}{2014}]%
        {zhang2014optimal}
\bibfield{author}{\bibinfo{person}{Weinan Zhang}, \bibinfo{person}{Shuai Yuan},
  {and} \bibinfo{person}{Jun Wang}.} \bibinfo{year}{2014}\natexlab{}.
\newblock \showarticletitle{Optimal real-time bidding for display advertising}.
  In \bibinfo{booktitle}{\emph{SIGKDD}}. \bibinfo{pages}{1077--1086}.
\newblock


\bibitem[\protect\citeauthoryear{Zhao, Qiu, Guan, Zhao, and He}{Zhao
  et~al\mbox{.}}{2018}]%
        {zhao2018deep}
\bibfield{author}{\bibinfo{person}{Jun Zhao}, \bibinfo{person}{Guang Qiu},
  \bibinfo{person}{Ziyu Guan}, \bibinfo{person}{Wei Zhao}, {and}
  \bibinfo{person}{Xiaofei He}.} \bibinfo{year}{2018}\natexlab{}.
\newblock \showarticletitle{Deep reinforcement learning for sponsored search
  real-time bidding}. In \bibinfo{booktitle}{\emph{SIGKDD}}.
  \bibinfo{pages}{1021--1030}.
\newblock


\bibitem[\protect\citeauthoryear{Zhou, Mou, Fan, Pi, Bian, Zhou, Zhu, and
  Gai}{Zhou et~al\mbox{.}}{2019}]%
        {zhou2019deep}
\bibfield{author}{\bibinfo{person}{Guorui Zhou}, \bibinfo{person}{Na Mou},
  \bibinfo{person}{Ying Fan}, \bibinfo{person}{Qi Pi}, \bibinfo{person}{Weijie
  Bian}, \bibinfo{person}{Chang Zhou}, \bibinfo{person}{Xiaoqiang Zhu}, {and}
  \bibinfo{person}{Kun Gai}.} \bibinfo{year}{2019}\natexlab{}.
\newblock \showarticletitle{Deep interest evolution network for click-through
  rate prediction}. In \bibinfo{booktitle}{\emph{AAAI}},
  Vol.~\bibinfo{volume}{33}. \bibinfo{pages}{5941--5948}.
\newblock


\bibitem[\protect\citeauthoryear{Zhou, Zhu, Song, Fan, Zhu, Ma, Yan, Jin, Li,
  and Gai}{Zhou et~al\mbox{.}}{2018}]%
        {zhou2018deep}
\bibfield{author}{\bibinfo{person}{Guorui Zhou}, \bibinfo{person}{Xiaoqiang
  Zhu}, \bibinfo{person}{Chenru Song}, \bibinfo{person}{Ying Fan},
  \bibinfo{person}{Han Zhu}, \bibinfo{person}{Xiao Ma},
  \bibinfo{person}{Yanghui Yan}, \bibinfo{person}{Junqi Jin},
  \bibinfo{person}{Han Li}, {and} \bibinfo{person}{Kun Gai}.}
  \bibinfo{year}{2018}\natexlab{}.
\newblock \showarticletitle{Deep interest network for click-through rate
  prediction}. In \bibinfo{booktitle}{\emph{SIGKDD}}.
  \bibinfo{pages}{1059--1068}.
\newblock


\bibitem[\protect\citeauthoryear{Zhu, Jin, Tan, Pan, Zeng, Li, and Gai}{Zhu
  et~al\mbox{.}}{2017}]%
        {zhu2017optimized}
\bibfield{author}{\bibinfo{person}{Han Zhu}, \bibinfo{person}{Junqi Jin},
  \bibinfo{person}{Chang Tan}, \bibinfo{person}{Fei Pan},
  \bibinfo{person}{Yifan Zeng}, \bibinfo{person}{Han Li}, {and}
  \bibinfo{person}{Kun Gai}.} \bibinfo{year}{2017}\natexlab{}.
\newblock \showarticletitle{Optimized cost per click in taobao display
  advertising}. In \bibinfo{booktitle}{\emph{SIGKDD}}.
  \bibinfo{pages}{2191--2200}.
\newblock


\bibitem[\protect\citeauthoryear{Zitzler and Thiele}{Zitzler and
  Thiele}{1998}]%
        {zitzler1998multiobjective}
\bibfield{author}{\bibinfo{person}{Eckart Zitzler} {and}
  \bibinfo{person}{Lothar Thiele}.} \bibinfo{year}{1998}\natexlab{}.
\newblock \showarticletitle{Multiobjective optimization using evolutionary
  algorithms—a comparative case study}. In \bibinfo{booktitle}{\emph{PPSN}}.
  Springer, \bibinfo{pages}{292--301}.
\newblock


\end{thebibliography}
\newpage 
 \appendix
\section*{Appendices}
\section{REPRODUCIBILITY}
We will provide details of our implementation and experimental setup to help reproduce the findings in this work. Though we cannot release the codes and datasets due to business secret and privacy issues, the proposed models are rather standard without sophisticated techniques. Hence, we believe it is easy to re-implement them with the information here.
\section{Distributed Coordinated MACG Algorithm}
In this section, we summarize the distributed coordinated MACG. We distribute billions of auction data into the cluster platform, and let each CPU process partial data. The details of the distributed coordinated MACG are shown in Algorithm \ref{DC-MACG}.

\begin{breakablealgorithm}
\makeatletter  
\caption{Distributed Coordinated MACG}
\makeatother
\begin{algorithmic}[1]
\label{DC-MACG}
\REQUIRE ~~\\
Learning rate $\alpha$,\\
Noise standard deviation $\delta$,\\
The numbers of multi-workers $W$,\\ 
Times of perturbations in each epoch $H$,\\ 
Total numbers of auctions $|J|$.\\
\STATE Initialization: initialize parameters population $\{\theta_{1w}\}_{w=1}^{w=W}$ randomly and send $J/W$ auctions in order from $\mathcal{J}$ for each worker.
\FOR{episodic for p = 1,2, ..., P}
\FOR{parameter w = 1,2,...,$W$}
\FOR{parameter h = 1,2,...,$H/W$}
\STATE Sample $\epsilon_{h} \sim N(0,I)$.
\STATE Update $\theta_{ph}$ =  $\theta_{pw} + \delta * \epsilon_{h}$.
\ENDFOR
\ENDFOR
\STATE Send the parameters set $\{\theta_{ph}\}_{h=1}^{H}$ to each worker
\FOR{multiple workers w = 1,2, ..., $W$}
\FOR{each worker j = 1, 2, ..., $|\mathcal{J}|$/W}
\STATE Simulate $j$th auction with parameters and save scalar $e_{ij}$, $w_{ij}$,$CTR_{ij}$ for winning AD $i$.
\ENDFOR
\ENDFOR
\STATE Send all scalars $e_{ij}$, $w_{ij}$,$CTR_{ij}$ from each worker to every other workers.
\STATE Compute ${M^{(1)}}$ $M^{(2)}$,$M^{(3)}$ and $M^{(0)}$ corresponding to policy parameters through Eq.~(\ref{click}), (\ref{GMV_AD}),(\ref{CART}),  (\ref{GMV}).
\STATE Compute ${M_{all}}$ corresponding to policy parameters through Eq.~(\ref{M_AD}),(\ref{M_all}) .
\STATE Select top W $\theta_{pw}$ through ${M_{all}}$ from parameters set $\{\theta_{ph}\}_{h=1}^{H}$ when $M^{(0)}$ and $M_{AD}$ are both greater than last epoch.
\STATE Save the seed parameters $\{\theta_{(p+1)w}\}_{w=1}^{w=W}$ for next epoch.
\ENDFOR

\end{algorithmic}
\end{breakablealgorithm}

\section{Implementation Details}
\subsection{Offline Simulator}
For the offline simulator construction, firstly, we need to record the complete online bidding information for each auction, including all ADs in the bidding queue, and related model-based features, such as CTR and CVR. These features do not change in the offline environment. We calculate the multi-objective scores of historical bid strategy as a benchmark. For simulated re-bidding of all auctions, the new bids are ordered by eCPM. Top AD wins the auction and the cost is made according to GSP. And then, we calculate the multi-objective scores caused by the new bids and evaluate the bidding strategy. The offline simulator essentially re-orders the ADs in the bidding queue by re-bidding. Alibaba platform uses OCPC strategy as the online bidding policy, so the multi-objective scores of OCPC are employed as the benchmark of offline simulator. 
\subsection{Details of Features}
To help reproduce the experiments, we provide detailed data characteristics. The features of AD $i$ under auction $j$ is defined as $\mathbf{S_{ij}}$. It contains related real-time contextual features: $CTR_{ij}$, $CVR_{ij}$, $IP_{ij}$, $WCVR_{ij}$, ${GMV_{ij}}$, ${COST_{ij}}$, $PPC_{ij}$ and $tk_{j}$. $\mathbf{S_{.j}}$ denotes the average of $\{\mathbf{S_{ij}}\}_{i \in \mathcal{I}_j}$.

\subsection{Hyper-parameter Settings}
Tab. \ref{t1} exhibits the settings of key hyper-parameters. For each agent, we use the same network structure with agent-specific parameters. 

\begin{table}
\center
\center \caption{Hyperparameter settings}\label{t1}
\begin{tabular}{c|c}
 \hline
   Hyper-parameter & Setting  \\
  \hline
$Agent^{(k)}$ Net & [8,4,1]\\
\hline
Shared Net & [8,4,1] \\
\hline
Allocation Net & [8,4,1] \\
\hline
H & 10000 \\
\hline
Learning Rate & 0.001 \\
\hline
Number of Workers & 2000 \\
\hline
Range & 0.3 \\
\hline
$\lambda_M$ & 1.2 \\
\hline
\end{tabular}
\end{table}

\subsection{Details of Policy Net}
As show in Tab.\ref{t1}, the $Agent^{(k)}$ Net, the Shared Net and the Allocation Net are three-layer neural networks with the same network architecture. The network setting [8,4,1] denotes the dimensionality of the input layer, hidden layer and output layer. We adopt the sigmoid function as the activation function for all the networks. For $Agent^{(k)}$ Net and Shared Net, we transform their outputs $x$ into [1-$range$, 1+$range$] interval through [$1 - range * (2 * 1 / (1 + e^{-x}) - 1)$,  $1 + range * (2 * 1 / (1 + e^{-x}) - 1)$]. The total number of parameters in the policy network is 180 ($180 = 5*(8*4+4)$). 

\subsection{Training Strategy}
Based on the offline simulator, we first initialize $H$ groups of parameters  $\{\theta_{ph}\}_{h=1}^{H}$ for the first epoch $p=1$. Then we collect the predicted objective values throughout the epoch and use the multi-objective score $M_{all}$ to select top $W$ groups of parameters as the seed parameters for the next epoch. We perform random perturbation (i.e., $\epsilon_{h} \sim N(0,I)$ where $\epsilon_{h}$ is a perturbation parameter) to generate a new population $\{\theta_{(p+1)h}\}_{h=1}^{H}$ of parameter vectors for the next iteration and simulate the above bidding process until all the objectives converge.


\subsection{Hardware and Software}
We deploy MACG on a cluster with 2000 CPUs and TensorFlowRS (TFRS) platform provided by Alibaba. TFRS is a distributed deep learning platform based on TensorFlow 1.7 used internally in Alibaba. In our experiments, the trainable dataset is distributed on 2000 workers (20 CPU cores for each worker), compute corresponding objective scores asynchronously and accumulate the objective score of each worker to update the model parameters.

\end{document}